\documentclass[10pt,twocolumn,a4paper]{esaAI}
\usepackage{graphicx}
\usepackage{subfigure}
\usepackage{url}
\usepackage{enumitem}
\newcommand{\gomez}{G\'{o}mez}

\newcommand{\etal}{\textit{et al.}}
\newcommand{\todo}[1]{\iffalse #1 \fi}
\setlist[enumerate]{itemsep=0mm}
\usepackage{microtype}
\microtypecontext{spacing=nonfrench}

\title{Delving into the Utilisation of ChatGPT in Scientific Publications in Astronomy}

\def\authorEmail{thesimoneastarita@gmail.com}

\author[1]{Simone Astarita\thanks{Corresponding author. E-Mail: \authorEmail}}
\author[1]{Sandor Kruk}
\author[1]{Jan Reerink}
\author[1]{Pablo \gomez}

\affil[1]{European Space Agency (ESA), European Space Astronomy Centre (ESAC), Camino Bajo del Castillo s/n, 28692 Villanueva de la Ca\~nada, Madrid, Spain}

\begin{document}

\makeCustomtitle

\begin{abstract}
Rapid progress in machine learning approaches to natural language processing has culminated in the rise of large language models over the last two years. Recent works have shown their unprecedented adoption in academic writing, but their pervasiveness in astronomy has not been studied sufficiently.
To remedy this, we extract words that ChatGPT uses more often than humans when generating academic text and search a total of 1 million articles for them. This way, we assess the frequency of word occurrence in published works in astronomy tracked by the NASA Astrophysics Data System since 2000. We then perform a statistical analysis of the occurrences.
We identify a list of words favoured by ChatGPT and find a statistically significant increase for these words against a control group in 2024, which matches the trend in other disciplines.
These results suggest a widespread adoption of these models in the writing of astronomy papers. We encourage organisations, publishers, and researchers to work together to establish ethical and pragmatic guidelines to maximise the benefits of these systems while maintaining scientific rigour.
\end{abstract}

\section{Introduction} \label{introduction}
The availability and sudden uptake of Large Language Models (LLMs) \cite{ReutersCGPT} raises questions for the scientific community: to what degree are they utilised in the creation of scientific publications? What effects, if any, does this technology have? How should the community e.g., journals that must set the conditions for an acceptable submission, respond?

Understanding of the process surrounding publications is important for public or international institutions that exist to advance science. Increasing availability of data, access to computing resources and popularity of applied machine learning in all steps of scientific knowledge production may alter the traditional process significantly. LLM-based writing tools can both impact a writer's expressive ability, which may be a factor in the peer-review process \cite{politzer2020preliminary}, as well as change the scientific quality of the content if the technology is used for more than its linguistic capabilities. While there is a risk of eroding writing skills, we may gain levelled playing field with respect to an international community attempting to publish in a non-native language.

Advances in machine learning have facilitated scaling up natural language processing models \cite{vaswani2017attention} to approach human capability in generating text \cite{brown2020language}.
Modern systems may leverage generative machine learning architectures with vast embedded data sets to generate answers to user inputs that correspond fairly well with user's expectations \cite{meyer2023chatgpt}, although the veracity of the output may not be guaranteed \cite{alkaissi2023artificial}. Some articles \cite{zamfiroiu2023chatgpt} discuss how scientists might use LLMs for writing papers, such as aiding in formulation and providing ideas, and others illustrate their risks, such as the questionable veracity of generated texts and the perceptions on whether generated text is equivalent to one's own work \cite{buriak2023best}. While scientific communities may react slowly to technological change \cite{kuhn1997structure}, surveys indicate a fast uptake of the technology \cite{adeshola2023opportunities}; however, despite interesting research on identification of generated text \cite{bhattacharjee2024fighting, sadasivan2023can}, reliable data on the rate of utilisation is scarce. While the detection of AI-generated passages is uncertain, it may be possible to detect utilisation at the population level: differences in linguistic features of generated texts relative to natural language have been used to quantify generated text in some scientific disciplines, showing, for example, that LLMs are more likely to use nouns and adjectives while relying on a smaller vocabulary \cite{liao2023differentiate}.

In the first quarter of 2024, several studies investigated the possible influence of ChatGPT on academic writing, either by looking at frequency changes in style words \cite{geng2024chatgpt}, or by leveraging their own GPT-human text corpus \cite{liang2024monitoring}, or by studying selected AI-favoured words in papers \cite{gray2024chatgpt}. They all concluded that ChatGPT usage was likely, but hard to estimate.

Following a prominent tweet by Nguyen on the use of the word \textit{delve} in PubMed articles \cite{nguyen2024tweet}, several studied the homonymous database \cite{sayers2022pubmed}. Some again observed frequency changes in some style words \cite{matsui2024pubmed1, masukume2024pubmed2}, while Kobak \etal{} leveraged excess word appearances to estimate AI usage \cite{kobak2024pubmed3}. Outside PubMed, Liu and Bu used AI-generated content detection tools to make similar assessments \cite{liu2024AIGC}, although the reliability of such tools has been challenged \cite{mitchell2023detectgpt}. Liang \etal{} created a corpus of human vs AI-generated academic text and used their previous framework \cite{liang2024monitoring} to assess the fraction of LLM-generated \cite{liang2024mapping}. While their predictions, hypotheses, and scientific field of inquiry vary, they all conclude that ChatGPT has heavily influenced academic writing in 2023 and especially 2024.

We contribute by investigating publications in the field of astronomy (see Section \ref{dataset}) by leveraging Liang \etal{}'s last dataset (see Section \ref{sec:we}) and then carrying out a simple yet effective linguistic analysis (see Section \ref{sec:wa}). We provide support for the hypothesis of widespread use of LLMs in astronomy publications (see Section \ref{results}) while highlighting limitations and potential for further research (see Section \ref{discussion}).


\section{Methods} \label{methods}

We identify words that appear overly frequently in AI-generated text when compared to human text, and assess their presence in astronomy papers over time. By AI, we here refer to the chat version of GPT \cite{brown2020language}, ChatGPT, the most widely adopted and accessible LLM.

Our code, data, and guidelines on how to reproduce these results are available on GitHub\footnote{\url{https://github.com/ESA-Datalabs/llm-usage-astronomy/}}.

\vspace{-0.1cm}
\subsection{Dataset} \label{dataset}

In order to estimate what words LLMs overuse, we utilise the data created by \cite{liang2024mapping}: a corpus produced by ChatGPT 3.5 (specifically, gpt-3.5-turbo-0125). The dataset was formed by taking an existing human-authored, pre-ChatGPT introduction, paragraph of an academic paper, making ChatGPT summarise it, and then requesting a new instance of ChatGPT to write the same paragraph based on the summary. We utilise the original, human-produced paragraphs to extract what words are ChatGPT-specific.

We are using NASA’s Astrophysics Data System (ADS) \cite{Accomazzi2015} API \cite{Lockhart2022} to identify words in the full text of astronomy papers on NASA ADS. 
The API returns the publications in which specific words appear, though it does not indicate the frequency of the words within each publication. To compare word usage over many years, we normalise by the number of papers published each year, since the annual publication count in astronomy has grown over time, particularly since the COVID-19 pandemic \cite{Bohmcovid}.

The dataset used in this study was obtained through the NASA ADS API on 24 May 2024. The dataset includes all the entries on NASA ADS between 1 January 2000 and 24 May 2024, in the ‘astronomy’ collection, which includes planetary science. There are 1\,061\,637 entries in this time interval of the type ‘article', of which 641\,656 are refereed and 419\,745 non-refereed. We include both refereed and non-refereed text indexed by NASA ADS, considering them relevant for the analysis as there may be differences in trends between them due to the time required for the refereeing process, and the process itself.


\subsection{Word Extraction} \label{sec:we}

We calculate the frequency of all tokens in both human and AI corpora created by Liang \etal{} \cite{liang2024mapping}, tokenized with the tokenizer of the large English model by spaCy \cite{spacy_2020}, after having removed numbers. We calculate the frequency for every possible token $t$ in the overall vocabulary $V$, excluding tokens that are not in both lists.

\begin{figure}[ht!]
\centering
\includegraphics[width=0.50\textwidth]{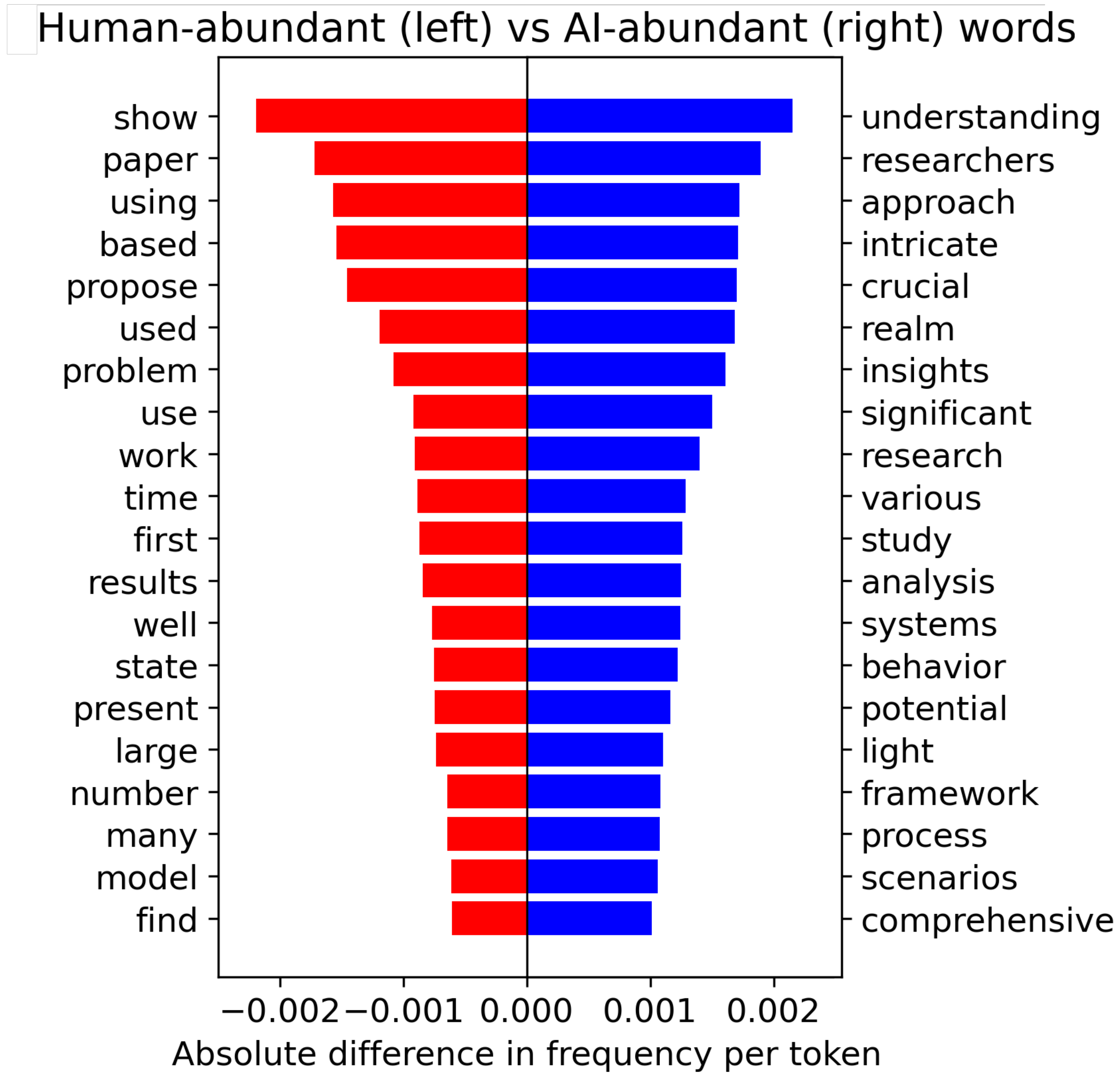}
\caption{Top content words with the highest \textit{absolute} difference of AI and human frequency (left) and vice-versa (right). The left column is not the same as the top AI-favoured words, which we calculate using \textit{relative} frequency change.}\label{aivshumtopabsolute}
\end{figure}

We calculate the ratio between the frequencies of words $w$ in the AI corpus $C_{\text{ai}}$ and the human corpus $C_{\text{hum}}$, and select the highest 100 values:
\begin{equation*}
    W = \text{Top}_{100}\left(t \in V \Bigg| \frac{|w \in C_{\text{ai}}|w = t|}{|C_{\text{ai}}|} : \frac{|{w \in C_{\text{hum}}|w = t}|}{|C_{\text{hum}}|}\right)
\end{equation*}

The words in \textit{W} show the highest changes in frequencies between AI-generated and human-written language. This word selection method does not rely on human judgement or frequency changes in 2023-24, so \textit{W} makes an ideal candidate for our analysis: a change in the usage of \textit{W}-words within the last two years may indeed suggest the use of ChatGPT.


Finally, we extract 100 random words as a control group $C$. We pick four random papers in NASA ADS from a random month for every year from 2000 to 2024. We then pick a random word in the abstract of these papers that has not been picked yet, is not in $W$ and is searchable by NASA ADS.

\subsection{Word Frequency Analysis} \label{sec:wa}

For each word in both $W$ and $C$, we use the keyword search function of NASA ADS to calculate how many publications between 2000 and 2024 contained that word, and then calculate the yearly frequency of occurrence in publications, regardless of how many times it appears in each publication.


We calculate the percentage difference in frequency of each word in $W$ from one year to the next. We then average these percentages for the top 5, 10, 20, 50 and for all of the words. Finally, we compare these numbers with the overall average for all the words in $C$.

\begin{figure}[ht!]
    \centering
    \begin{subfigure}
        \centering
        \includegraphics[trim={0.5cm 1cm 1.5cm 1.8cm}, clip, width=\linewidth]{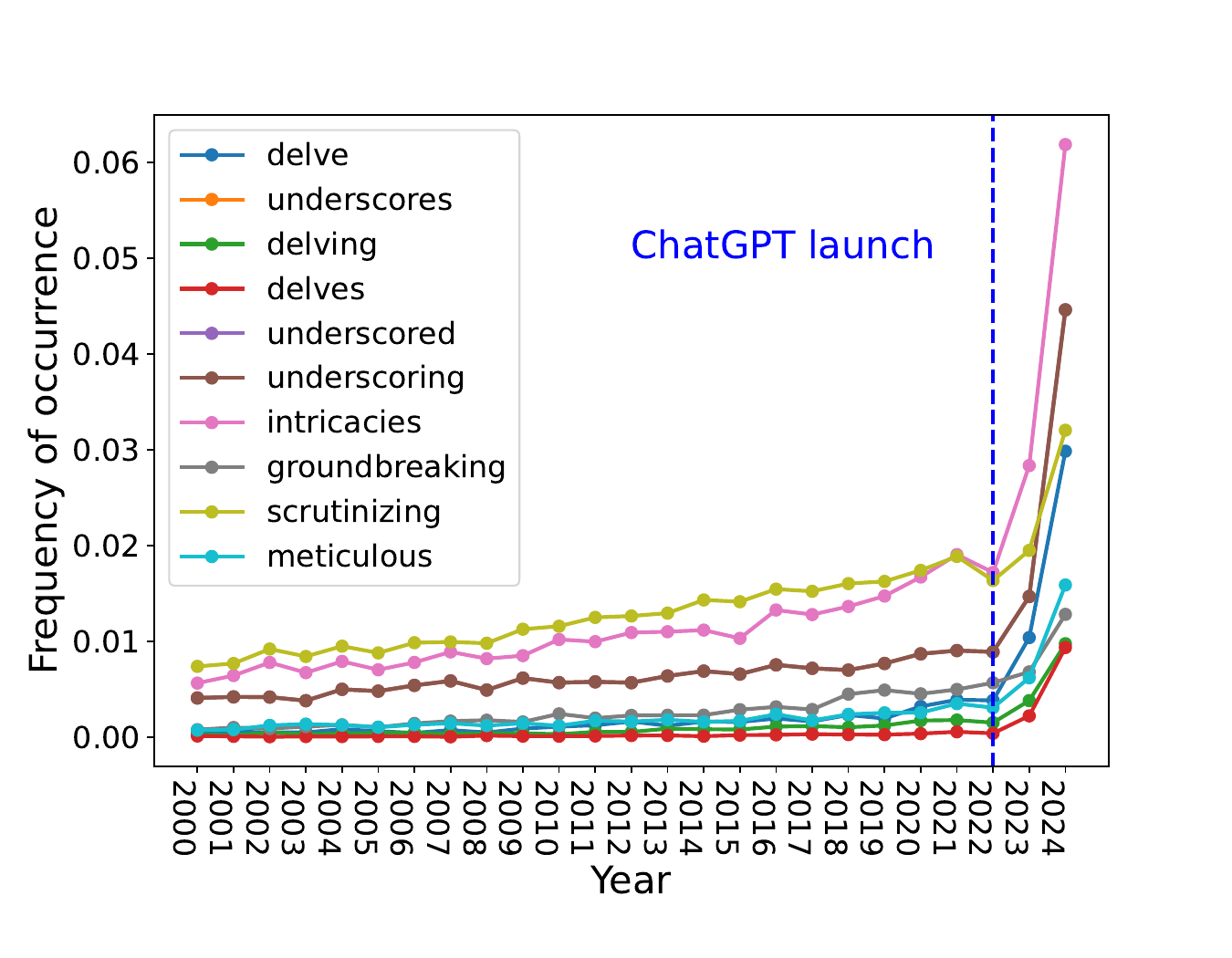}
        \caption{Frequency of the occurrence of the top 10 AI-favoured words in astronomy articles as a function of time. These 10 words are the ones with the highest relative frequency-per-token difference between AI and human text in the Liang \etal{} \cite{liang2024mapping} corpora.}
        \label{aivshumtop}
    \end{subfigure}
    \hfill
    \begin{subfigure}
        \centering
        \includegraphics[trim={0.5cm 1cm 1.5cm 1.4cm}, clip, width=\linewidth]{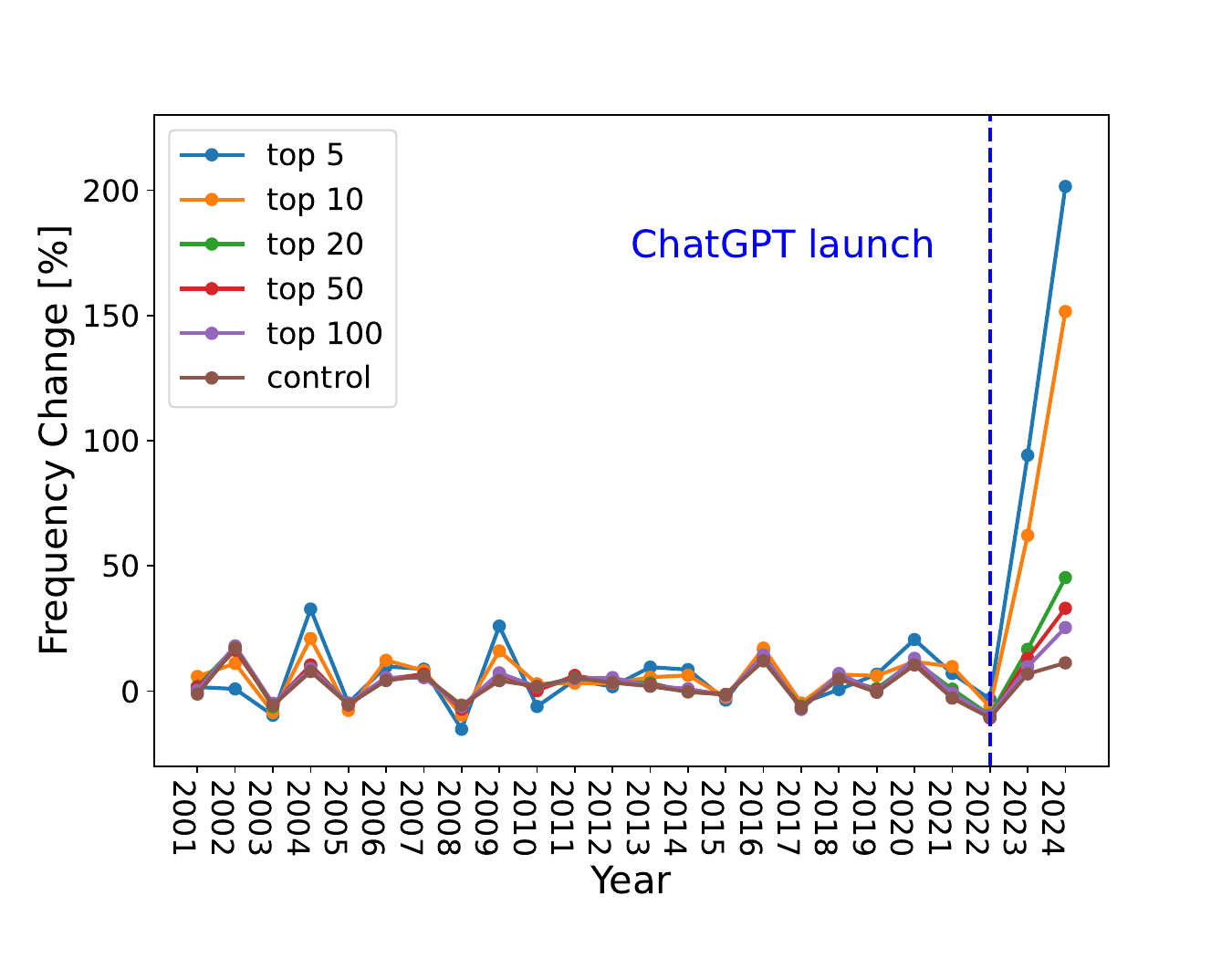}
        \caption{The relative frequency of occurrence change year by year (in percentages) for subsets of AI-favoured words appearing in all astronomy articles. These words are the ones with the highest relative frequency-per-token difference between AI and human text in the Liang \etal{} \cite{liang2024mapping} corpora. 
        }
        \label{perc_fr}
    \end{subfigure}
\end{figure}


To determine the significance of the changes, we apply the two-sample Kolmogorov–Smirnov test. We treat the frequency of occurrence as a random variable of the words in $W$ and compare the frequency of 2024 with each preceding year. If there is no significant change, we will see relatively high p-values, indicating that the underlying distribution remains the same. We use 2010 as the control year, to ensure that the selected words do not display the same behaviour every year.

Finally, we repeat this analysis for two complimentary subsets of our data, peer-reviewed works and not, to assesses whether the reviewing process influences the results.

\begin{figure}[ht!]
\centering
{\includegraphics[width=1.0\linewidth]{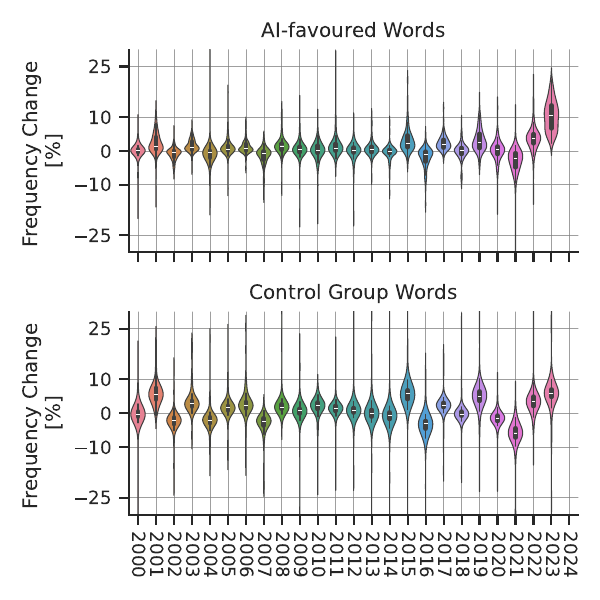}}
\caption{Change in frequency of occurrence of all 100 AI-favoured words and all 100 random words in the control sample as a function of time. These words are the ones with the highest relative frequency-per-token difference between AI and human text in the Liang \etal{} \cite{liang2024mapping} corpora.}\label{vplot}
\vspace*{-0.5cm}
\end{figure}

\section{Results} \label{results}

As shown in \cref{perc_fr}, the average percentage change in frequency for the top words in $W$ increases dramatically in 2023 and 2024, unlike the average word in the control sample $C$.  The further we move from the top words, the more the pattern in 2024 resembles that of the control group. In our results, this behaviour is even more accentuated in non-peer-reviewed works.

The Kolmogorov–Smirnov analysis between year 2024 and all other years for the words in $W$ and its subsets (top 5, 10, 20 and 50) results in all p-values under 0.05, with the exception of the year 2023\footnote{We refer to our GitHub repository for the complete set of p-values.}. This indicates that the word frequencies for 2024 are not extracted from the same distribution, suggesting that the 2024 frequencies are not just minor variations of the previous ones. Some variation is due to occur, as shown in \cref{vplot}, where we show the percentage of change from year to year in word frequency for $C$ and $W$, normalised by mean frequency change for each word. To rule out the possibility that this is merely a statistical fluctuation, we compared the results to a control group. Our analysis shows that p-values between 2024 and previous years up to 2016 are well above 0.05. So, while variation exists, it is significantly less pronounced for random words in years close to 2024. 

Finally, the words in $W$ may naturally display greater yearly variation, so that their frequency changes in 2024 are not correlated with the advent of ChatGPT. In order to exclude this possibility, we repeat the analysis for a control year (2010). The p-values for 2010 compared to each year between 2000 and 2020 are all well above 0.05 for the words in $W$ and its subsets, consistent with the results for the words in $C$ as well. This shows that such significant variation is not typical for these words.

Thus, we conclude that the frequency distribution of words in $W$ is significantly different in 2024 compared to all years before 2023, and this variation is unlikely due to random language fluctuations. This analysis, repeated for both peer-reviewed and non-peer-reviewed works, shows that the phenomenon is more pronounced in non-peer-reviewed works.

We caution that this present work does not estimate the amount of AI-generated text in published works, nor do we claim that the presence of words in $W$ indicates a paper was written with ChatGPT. Instead, we observe that the words in $W$, which are favoured by ChatGPT, appear in a much larger portion of papers in 2024 compared to previous years, a phenomenon not observed for other random words.

In short, our study provides evidence that:
\begin{itemize}[itemsep=0pt, parsep=0pt, topsep=2pt]
    \item ChatGPT-generated text overuses the words in $W$, such as \textit{delve} and \textit{underscore}, when compared to humans.
    \item In astronomical publications, the frequency of occurrence of words in $W$ increased significantly in 2024, especially for the words near the top, and does so to a higher extent when compared to other control words ($C$) or years.
    \item This phenomenon is more pronounced in non-peer-reviewed works.
\end{itemize}
\section{Discussion} \label{discussion}
The adoption of LLMs in scientific writing brings both chances and challenges with it. Amid vivid discussion in academia \cite{fecher2023friend,meyer2023chatgpt,rahman2023}, we hope to spark this conversation in the astronomical community as well.

\subsection{Limitations}
The presented data and analysis have some limitations: a study of the temporal evolution of scientific writing is faced with many confounders. We were unable to control for changes due to differences in the distribution of authors' native languages, which are known to impact language \cite{carson1990reading}. Our analysis does not directly consider the average paper length changes over time, which can impact the frequency of occurrence; however, we hope to have mitigated this effect by comparing our results with a control group of words. We also considered only whether a word is present or not in the text, not the number of times it appears. We controlled for language change caused by other factors \cite{labov2011principles} by investigating the change before the release of ChatGPT in November 2022. Finally, our scope is limited to identifying ChatGPT-related characteristics given its high adoption rate; the impact of other LLMs is hence neglected.

\subsection{Academic Implications}
At the current rate, it seems unavoidable that the proliferation of LLMs as tools in scientific writing will continue to increase given their utility. The initial reaction to the discovery of AI-written content in scientific articles and reviews has been mixed and publishers and academic actors are facing challenges in formulating guidelines \cite{fecher2023friend,Koplin2023-KOPPAE,porsdam2023generative}. Current approaches, e.g. by IEEE \footnote{\url{https://journals.ieeeauthorcenter.ieee.org/become-an-ieee-journal-author/publishing-ethics/guidelines-and-policies/submission-and-peer-review-policies/#ai-generated-text} Accessed: 2024-05-23} or MNRAS \footnote{\url{https://academic.oup.com/mnras/pages/general_instructions#Authorship} Accessed: 2024-05-23}, call for a disclosure of the utilisation in articles. Critics point out issues ranging from the risk of plagiarism and misrepresentation of facts to fabrication of references \cite{Koplin2023-KOPPAE}. Authors may also get less credit for the value of something created with AI while facing increased risk of backlash in case of errors or issues \cite{porsdam2023generative}. Further, there are some concerns that, if large amounts of data available in the future are AI-generated, models trained on that data may eventually deteriorate in quality \cite{shumailov2024curse}. Less attention has been given to prospective benefits of LLMs, such as helping non-native speakers of English, high-quality feedback for underprivileged researchers or a potential ease of the burden placed on scientists to review and stay up-to-date with an ever larger stream of new articles.

Many of these aspects are likely similar in astronomy and, as we can see in our analysis, this proliferation is unlikely to be ceasing. Additionally, with models continually improving, it might be hard to reliably detect their undisclosed usage or assess if LLMs misrepresent facts at a higher rate than humans \cite{sadasivan2023can}. In order to prevent a negative impact on scientific writing, we may want to improve LLMs e.g., with specialised models such as \textit{AstroLLaMA} 
\cite{nguyen2023,perkowski2024astrollama} or by allowing models to have up-to-date access to peer-reviewed scientific literature using Retrieval-Augmented Generation (RAG) approaches. \\


\textbf{Acknowledgements}. We acknowledge the extensive work behind the NASA Astrophysics Data System (\url{https://ui.adsabs.harvard.edu/}), an open-access publication repository that makes this and many other (meta-)research studies in the field possible. In particular, we thank Kelly Lockhart, who provided extensive information on the way the NADA ADS API works. We also thank Kartheik Iyer for the valuable discussions regarding this work and for providing information on accessing astronomy publications.

\printbibliography
\addcontentsline{toc}{section}{References}

\end{document}